\documentclass[journal]{IEEEtran}
\usepackage[margin=1.1in]{geometry} 
\pdfoutput=1

\usepackage{amsmath,amsfonts}
\usepackage{algorithmic}
\usepackage{algorithm}
\usepackage{array}
\usepackage[caption=false,font=normalsize,labelfont=sf,textfont=sf]{subfig}
\usepackage{textcomp}
\usepackage{stfloats}
\usepackage{url}
\usepackage{verbatim}
\usepackage{graphicx}
\usepackage{tcolorbox}
\usepackage{physics}
\usepackage{multirow}
\usepackage{multicol}
\usepackage{fontawesome5}
\usepackage{rotating}
\usepackage{cite}
\usepackage{booktabs}
\usepackage{float}
\usepackage{authblk}
\begin{document}

\title{Architectures of Topological Deep Learning:\\ A Survey of Message-Passing Topological Neural Networks}

\author{Mathilde Papillon$^{*1}$ \and Sophia Sanborn$^2$ \and Mustafa Hajij$^3$ \and Nina Miolane$^2$}
\affil{
$^1$Department of Physics \\ %
	$^2$Department of Electrical and Computer Engineering \\ 
    University of California, Santa Barbara \\ \vspace{0.25cm}
    $^3$ Department of Data Science \\
    University of San Francisco \\ \vspace{0.25cm}
    $^*$Corresponding Author: \texttt{papillon@ucsb.edu}
    \normalsize
}

\markboth{Journal of \LaTeX\ Class Files,~Vol.~14, No.~8, August~2021}%
{Shell \MakeLowercase{\textit{et al.}}: A Sample Article Using IEEEtran.cls for IEEE Journals}

\IEEEpubid{0000--0000/00\$00.00~\copyright~2023 IEEE}

\maketitle

\begin{abstract}
The natural world is full of complex systems characterized by intricate relations between their components: from social interactions between individuals in a social network to electrostatic interactions between atoms in a protein. Topological Deep Learning (TDL) provides a comprehensive framework to process and extract knowledge from data associated with these systems, such as predicting the social community to which an individual belongs or predicting whether a protein can be a reasonable target for drug development. TDL has demonstrated theoretical and practical advantages that hold the promise of breaking ground in the applied sciences and beyond. However, the rapid growth of the TDL literature for relational systems has also led to a lack of unification in notation and language across message-passing Topological Neural Network (TNN) architectures. This presents a real obstacle for building upon existing works and for deploying message-passing TNNs to new real-world problems. To address this issue, we provide an accessible introduction to TDL for relational systems, and compare the recently published message-passing TNNs using a unified mathematical and graphical notation. Through an intuitive and critical review of the emerging field of TDL, we extract valuable insights into current challenges and exciting opportunities for future development.
\end{abstract}

\begin{IEEEkeywords}
Deep learning, topology, message passing, graph, hypergraph, simplicial complex, cellular complex, combinatorial complex
\end{IEEEkeywords}

\section{Introduction}\label{sec1}
\IEEEPARstart{M}{any} natural systems as diverse as social networks \cite{knoke2019social} and proteins \cite{jha2022prediction} are characterized by \textit{relational structure}. This is the structure of interactions between components in the system, such as social interactions between individuals or electrostatic interactions between atoms.
In Geometric Deep Learning \cite{bronstein2021geometric}, Graph Neural Networks (GNNs) \cite{zhou2020graph} have demonstrated remarkable achievements in processing relational data using graphs\textemdash mathematical objects commonly used to encode \textit{pairwise relations}.

However, the pairwise structure of graphs is limiting. Social interactions can involve more than two individuals, and electrostatic interactions more than two atoms. Topological Deep Learning (TDL) \cite{hajij2023tdl,bodnar2023thesis} leverages more general abstractions to process data with higher-order relational structure. The theoretical guarantees \cite{bodnar2021topological,bodnar2021weisfeiler,huang2021unignn} of its models, Topological Neural Networks (TNNs),  lead to state-of-the-art performance on many machine learning tasks \cite{dong2020hnhn,hajij2022higher,barbarossasardellitti2020topological,chen2022bscnets}\textemdash and reveal high potential for the applied sciences and beyond. 

However, the abstraction and fragmentation of mathematical notation across the TDL literature significantly limits the field's accessibility, while complicating model comparison and obscuring opportunities for innovation. 
To address this, we present an intuitive and systematic comparison of published message-passing TNN architectures, heretofore referred to as TNNs. We contribute:
\IEEEpubidadjcol
\begin{itemize}
    \item \textbf{A pedagogical resource} accessible to newcomers interested in applying TNNs to real-world problems.
    \item \textbf{A comprehensive and critical review} of TNNs, their implementations and practical applications, with equations rewritten in our notation available at \url{https://github.com/awesome-tnns}.
    \item \textbf{A summary of open research questions}, challenges, and opportunities for innovation.
\end{itemize}

By establishing a common and accessible language in the field, we hope to provide newcomers and experienced practitioners alike with a solid foundation for cutting-edge research in TDL.

Other literature reviews at the intersection of topology and machine learning have focused on data representation~\cite{torres2021why} and physics-inspired models~\cite{battiston2021physics}. Message-passing TNNs are part of a broader spectrum of machine learning architectures leveraging topology. First surveyed in~\cite{hensel2021survey}, this spectrum encompasses additional methods such as topological data analysis for machine learning. In such cases, features computed from techniques such as persistent homology are used to enhance data representation or model selection. 

 \section{Topological Neural Networks}\label{sec3}
\begin{figure*}[!ht]
    \centering
    \includegraphics[width=\textwidth]{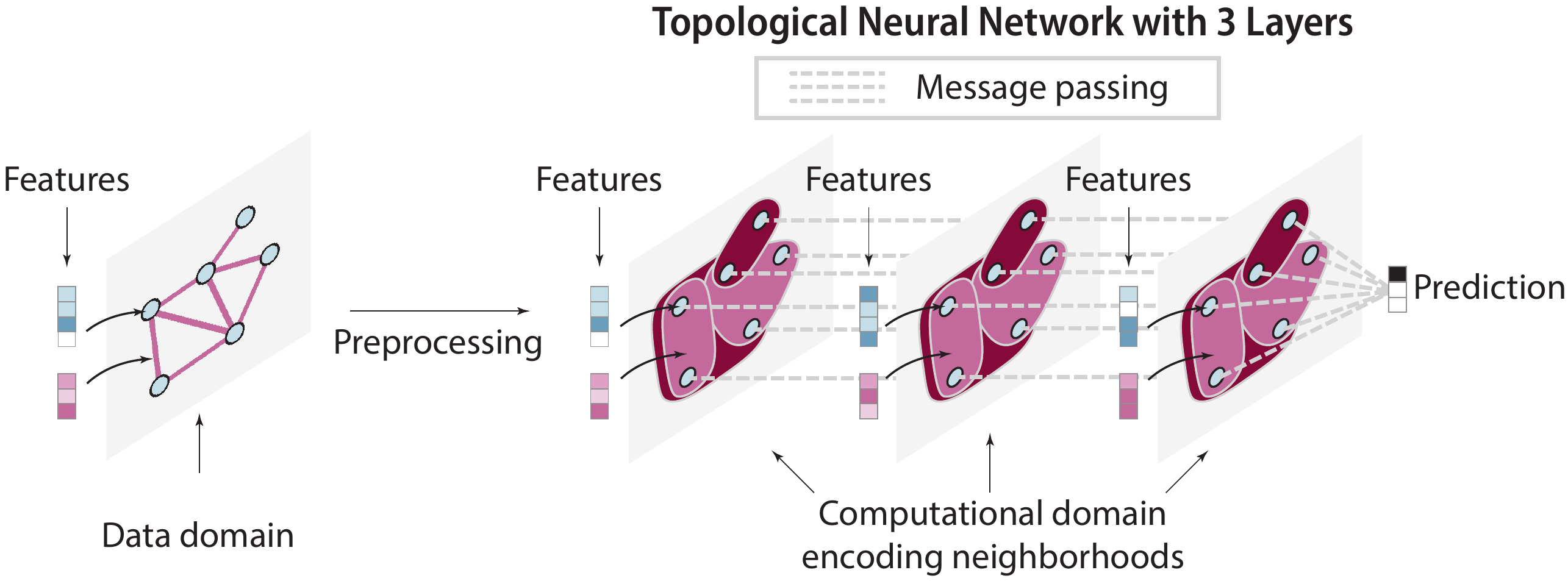}
    \caption{\textbf{Topological Neural Network}: Data associated with a complex system are features defined on a \textit{data domain}, which is preprocessed into a \textit{computational domain} that encodes interactions between the system's components with \textit{neighborhoods}. The TNN's layers use \textit{message passing} to successively update features and yield an output, e.g. a categorical label in classification or a quantitative value in regression. The output represents new knowledge extracted from the input data.}
	\label{fig:tnn}
\end{figure*}

Topological Neural Networks (TNNs) are deep learning architectures that extract knowledge from data associated with topologically rich systems such as protein structures, city traffic maps, or citation networks. A TNN, like a GNN, is comprised of stacked layers that transform data into a series of features (Figure \ref{fig:tnn}). Each layer leverages the fundamental concepts of \textit{data and computational domains}, \textit{neighborhoods}, and \textit{message passing}\textemdash presented in this section.

\subsection{Domains}

\begin{figure*}[!ht]
    \centering
    \includegraphics[width=0.8 \textwidth]{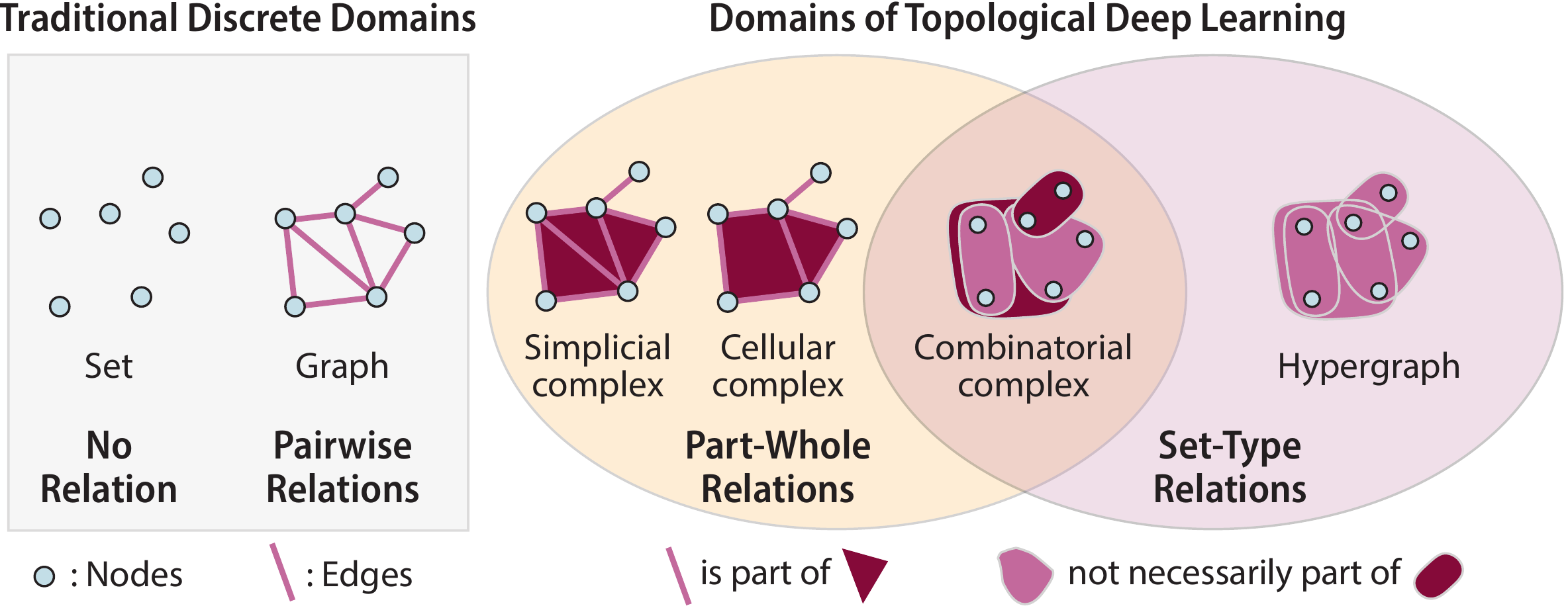}
	\caption{\textbf{Domains:} Nodes in blue, (hyper)edges in pink, and faces in dark red. Figure adapted from \cite{hajij2022higher}.}
	\label{fig:domains}
\end{figure*}

In Topological Deep Learning (TDL), data are features defined on discrete domains \cite{hajij2023tdl,bodnar2023thesis}. Traditional examples of discrete domains include sets and graphs (Figure \ref{fig:domains}, Left). A \textbf{set} is a collection of points called \textit{nodes} without any additional structure. A \textbf{graph} is a set with \textit{edges} that encode \textit{pairwise relations} between nodes, representing either geometric proximity or more abstract relationships. For example, a graph may represent a protein, with nodes encoding its atoms and edges encoding the pairwise bonds between them. Alternatively, a graph may represent a social network, where nodes represent individuals and edges denote social relationships. The domains of TDL generalize the pairwise relations of graphs to \textit{part-whole} and \textit{set-types} relations that permit the representation of more complex relational structure (Figure \ref{fig:domains}, Right) \cite{hajij2023tdl}. Here, we describe the key attributes of each domain and highlight their suitability for different data types. We refer the reader to \cite{torres2021why} and \cite{hajij2023tdl} for more extensive discussions. 




\begin{tcolorbox}

\begin{center}\normalsize\textbf{Beyond Graphs: The Domains of Topological Deep Learning}\end{center} \vspace{0.25cm} 
\small
\textsc{\underline{Set + Pairwise Relations}}\\ 

\textbf{Graph}:  A set of points (\textit{nodes}) connected with \textit{edges} that denote pairwise relationships. \\

\textsc{\underline{Set + Part-Whole Relations}}\vspace{.25cm}

\textbf{Simplicial Complex} (SC): A generalization of a graph in which three edges can form a \textit{triangular face}, four triangles can form a \textit{tetrahedral volume}, and so on. Edges only connect pairs of nodes. \\ 




\textbf{Cellular Complex} (CC):  A generalization of an SC in which \textit{faces, volumes, etc} are not restricted to be triangles or tetrahedrons but may instead take any shape. Still, edges only connect pairs of nodes. \\

\textsc{\underline{Set + Set-Type Relations}}\vspace{.25cm}

\textbf{Hypergraph} (HG):  A generalization of a graph, in which higher-order edges called \textit{hyperedges} can connect arbitrary \textit{sets} of two or more nodes. \\


\textsc{\underline{Set + Part-Whole and Set-Type Relations}}\vspace{.25cm}

\textbf{Combinatorial Complex} (CCC): A structure that combines features of HGs and CCs. Like an HG, edges may connect any number of nodes. Like a CC, cells can be combined to form higher-ranked structures. 


\end{tcolorbox}

\begin{figure*}[!ht]
    \centering
    \includegraphics[width=\textwidth]{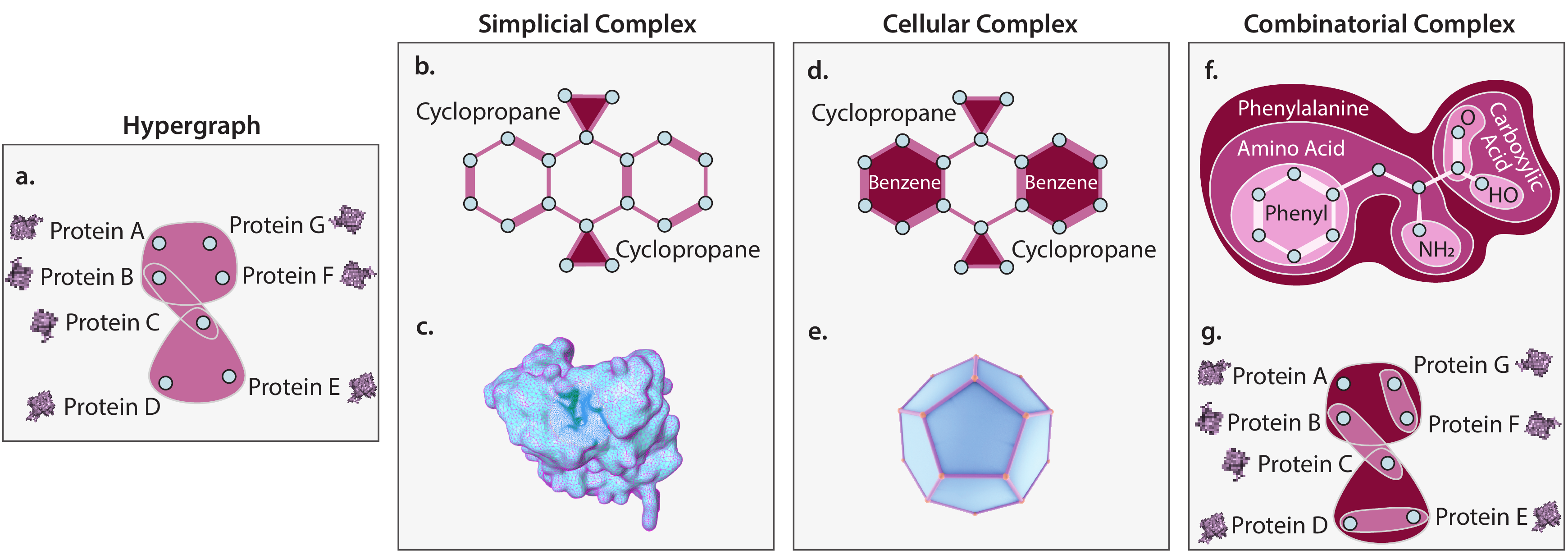}
    \caption{\textbf{Examples of Data on Topological Domains.} (a) Higher-order interactions in protein networks. (b) Limited molecular representation: rings can only contain three atoms. (c) Triangular mesh of a protein surface. (d) More flexible molecular representation, permitting the representation of any ring-shaped functional group. (e) Flexible mesh which includes arbitrarily shaped faces. (f) Fully flexible molecular representation, permitting the representation of the complex nested hierarchical structure characteristic of molecules and other natural systems. (g) Hierarchical higher-order interactions in protein networks.}
    \label{fig:domain-applications}
\end{figure*}

\textbf{Simplicial complexes (SCs)} generalize graphs to incorporate hierarchical \textit{part-whole relations} through the multi-scale construction of \textit{cells}. Nodes are \textit{rank 0} cells that can be combined to form edges (\textit{rank 1} cells). Edges are, in turn, combined to form faces (\textit{rank 2} cells), which are combined to form volumes (\textit{rank 3} cells), and so on. As such, an SC's faces must be triangles, volumes must be tetrahedrons, and so forth. SCs are commonly used to encode discrete representations of 3D geometric surfaces represented with triangular meshes (Figure \ref{fig:domain-applications}). They may also be used to represent more abstract relations; however, there is a risk of introducing spurious connections if the strict geometric constraints of an SC are not respected by the data\textemdash a point we elaborate on in Section \ref{section:limitation}.



\textbf{Cellular complexes (CCs)} generalize SCs such that cells are not limited to simplexes: faces can involve more than three nodes, volumes more than four faces, and so on. This flexibility endows CCs with greater expressivity than SCs \cite{bodnar2021weisfeiler}. A practitioner should consider employing this domain when studying a system that features part-whole interactions between more than three nodes, such as a molecule with benzene rings (Figure \ref{fig:domain-applications}).






\textbf{Hypergraphs (HGs)} extend graphs in that their edges, called \textit{hyperedges}, can connect more than two nodes. Connections in HGs represent \textit{set-type relationships}, in which participation in an interaction is not implied by any other relation in the system. This makes HGs an ideal choice for data with abstract and arbitrarily large interactions of equal importance, such as semantic text and citation networks. Protein interaction networks (Figure \ref{fig:domain-applications}) also exhibit this property: an interaction between proteins requires a precise set of molecules\textemdash no more and no less. The interaction of Proteins A, B, and C does not imply an interaction between A and B on their own. 


 
\textbf{Combinatorial complexes (CCCs)} generalize CCs and HGs to incorporate both \textit{part-whole} and \textit{set-type} relationships \cite{hajij2022higher,hajij2023tdl}. The benefit of this can be observed in the example of molecular representation. The strict geometric constraints of simplicial and cellular complexes are too rigid for capturing much of hierarchical structure observed in molecules. By contrast, the flexible but hierarchically ranked hyperedges of a combinatorial complex can capture the full richness of molecular structure, as depicted in Figure \ref{fig:domain-applications}. This is the most recent and most general topological domain, introduced in 2022 by \cite{hajij2022higher} and further theoretically established in \cite{hajij2023tdl}.




\subsubsection{Terminology}

Across discrete domains, we use the term \textit{cell} to denote any node or relation between nodes such as (hyper)edges, faces, or volumes. Cells possess two attributes: \textit{size}\textemdash the number of cells it contains\textemdash and \textit{rank}\textemdash where nodes are said to have rank 0, edges and hyperedges rank 1, faces rank 2, and so on. The part-whole relationships of simplicial and cellular complexes impose a relationship between the \textit{rank} of a cell and its \textit{size}: cells of rank $r$ contains exactly (resp. at least) $r + 1$ cells of rank $r - 1$: faces ($r = 2$) contain exactly (resp. at least) three edges ($r - 1 = 1$). By contrast, hypergraph cells do not encode part-whole relations and hyperedges may have any size. However, hypergraph cells are limited to ranks 0 and 1. A combinatorial complex is unrestricted in both rank and size: nodes have rank 0 and cells of any size $>$ 1 can have any rank.

\begin{figure}[!ht]
    \centering
    \includegraphics[width=0.7 \linewidth]{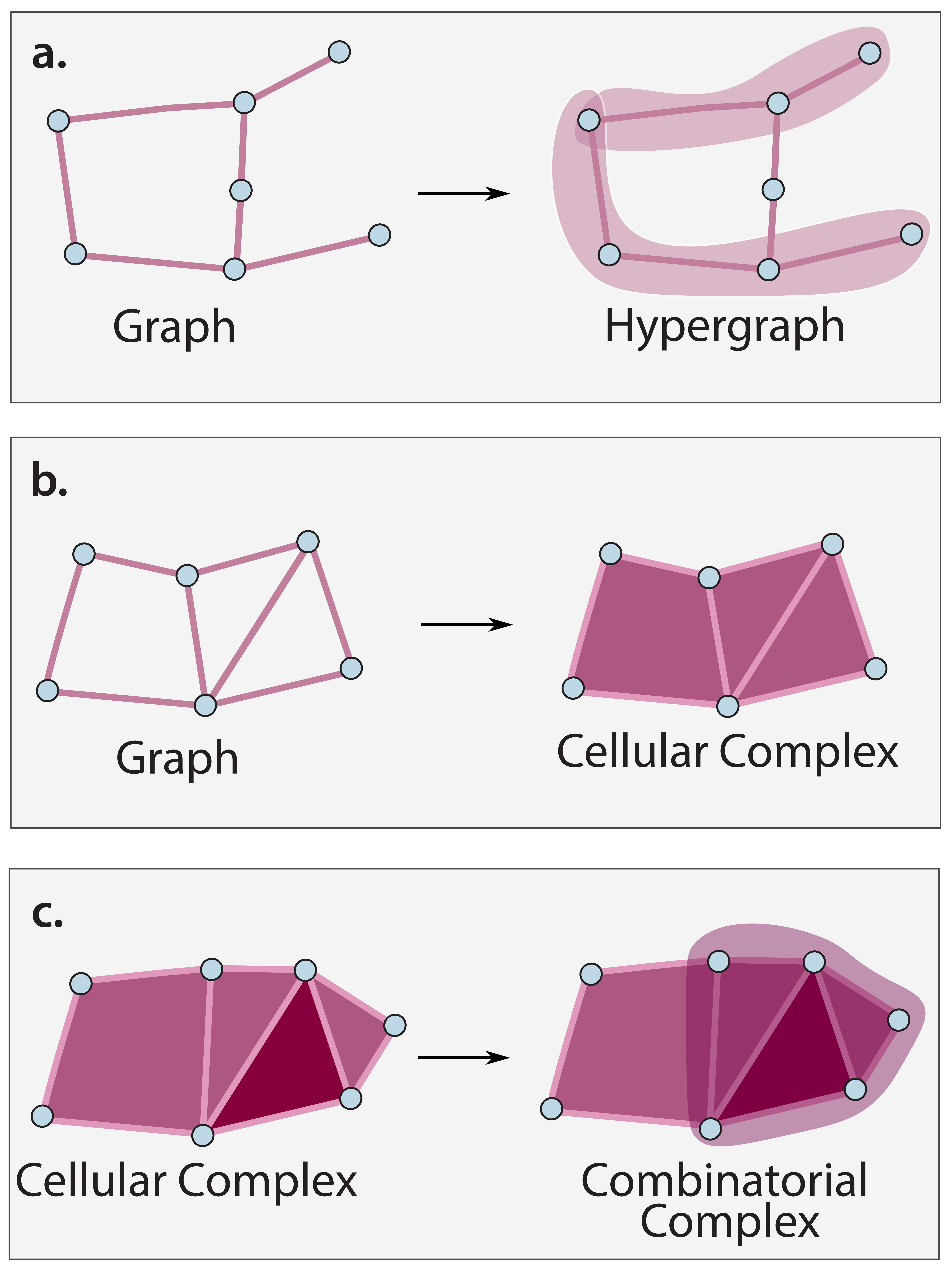}
    \caption{\textbf{Lifting Topological Domains.} (a) A graph is lifted to a hypergraph by adding hyperedges that connect groups of nodes. (b) A graph can be lifted to a cellular complex by adding faces of any shape. (c) Hyperedges can be added to a cellular complex to lift the structure to a combinatorial complex. Figure adopted from \cite{hajij2023tdl}.}
    \label{fig:lifting_maps}
\end{figure}

There is an important distinction between the inherent domain of the data (the \textit{data domain}) and the domain in which the data will be processed within a TNN: the \textit{computational domain}. Data defined on a graph, for example, may be ``lifted'' (Figure \ref{fig:lifting_maps}) to an alternative domain through a pre-processing stage (Figure \ref{fig:tnn}). For instance, a protein originally given as the graph of its atoms (nodes) and covalent bounds (edges) may be lifted into a CC computational domain that explicitly represents its rings (faces). In this review, \textit{domain} refers to the computational domain. Additionally, the computational domain may be \textit{dynamic}, changing from layer to layer in a TNN. 

\vspace{.25cm}

\begin{tcolorbox}
\begin{center}
    \textbf{Dynamic Domains}
\end{center}
\vspace{.25cm}
    \small
    
    \textbf{Static vs. Dynamic}: In a TNN, a \textit{static} domain is identical for each layer. For example, all three layers in Figure \ref{fig:tnn} operate on the same CCC, only features evolve across layers. A \textit{dynamic} domain changes from layer to layer. Nodes can be added or removed, edges can be rewired, and so on.
\end{tcolorbox}

\vspace{.25cm}

\subsubsection{Limitations}
\label{section:limitation}
An important limitation of both SCs and CCs is that faces (and analogous higher-order structures) can only form rings; the nodes on the boundary of the face must be connected in pairs. In many cases, this requirement is too stringent and can introduce artificial connections to the domain \cite{efficient2022yang}. For instance, lifting a citation network into an SC necessarily requires that any set of three co-authors having written a paper together (A, B, C) are also pairwise connected (A and B, A and C, B and C), even if no paper was ever exclusively authored by authors A and B, authors A and C, or authors B and C. \cite{efficient2022yang} propose a ``relaxed'' definition of the SC that remedies this. They show how training a TNN on such a modified domain increases performance. We note that even with artificial connections, SCs and CCs allow TNNs to leverage richer topological structure and avoid computational problems faced by GNNs \cite{rusch2023survey}. We further note that any topological domain is mathematically equivalent to a (possibly larger) graph \cite{velivckovic2022message}. We choose to express domains in their form above in order to provide better intuition to newcomers and reflect the widely adopted approaches in the literature.

\subsubsection{Features on a Domain}

Consider a domain, denoted $\mathcal{X}$, encoding relationships between components of a system. Data on the domain are represented as features supported on the domain's cells. Typically, features are vectors in $\mathbb{R}^d$ that encode attributes of each cell. For example, features may encode the atom (node), bond (edge), and functional group (face) types in a molecule. A feature associated with the interaction between a set of drugs (hyperedge) could indicate the probability of adverse reaction.

\begin{figure}[ht]
    \centering
    \includegraphics[width=0.9 \linewidth]{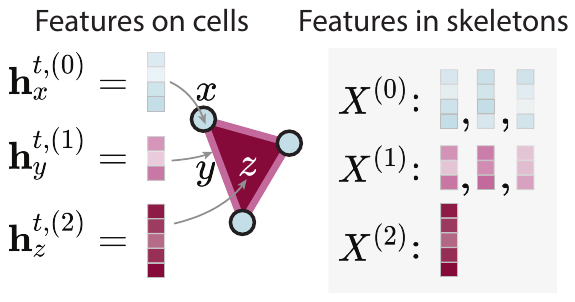}
	\caption{\textbf{Features on a Domain.} Left: Features onto three cells\textemdash$x$, $y$, and $z$. Right: Skeletons for the entire complex: $X^{(0)}$ contains node features, $X^{(1)}$ contains edge features, and so on. \vspace{-.25cm}}
	\label{fig:signal}
\end{figure}

We denote with $\mathbf{h}_x^{t, (r)}$ a feature supported on the cell $x \in \mathcal{X}$ at layer $t$ of the TNN, with $r$ indicating the rank of $x$ (Figure \ref{fig:signal}). The domain is decomposed into ranks, with $X^{(r)}$, or \textit{$r$-skeleton}, referring to all cells of rank $r$. Features can be categorical or quantitative. If the feature dimension varies across skeletons, the domain is \textit{heterogeneous}.

\begin{tcolorbox}
\begin{center}
    \textbf{Heterogeneous Domains}
\end{center}
\vspace{.25cm}
    \small

\textbf{Homogeneity vs. Heterogeneity}: In a \textit{heterogeneous} domain, the dimension $d_r$ of a feature $\mathbf{h}_x^{(r)}$ depends on the rank $r$ of the cell $x$ supporting it. A \textit{homogeneous} domain uses the same dimensionality $d$ for all ranks. 

\end{tcolorbox}

The features assigned to each cell may come directly from the data or be hand-designed by the practitioner. Alternatively, features can be assigned in a pre-processing stage using \textit{embedding} methods, which compute cell feature vectors that encode the local structure of the space. For graphs, methods such as DeepWalk \cite{perozzi2014deepwalk} and Node2Vec \cite{grover2016node2vec} are commonly used to embed nodes. Recent works have generalized these approaches to topological domains: Hyperedge2Vec \cite{sharma2018hyperedge2vec} and Deep Hyperedge \cite{payne2019deep} for hypergraphs, Simplex2Vec \cite{billings2019simplex2vec} and k-Simplex2Vec \cite{hacker2020k} for simplicial complexes, and Cell2Vec \cite{hajij2020cell} for cellular complexes.



\subsection{Neighborhood Structure}\label{neighborhood-structure}

\begin{figure}[!ht]
    \centering
    \includegraphics[width=0.8\linewidth]{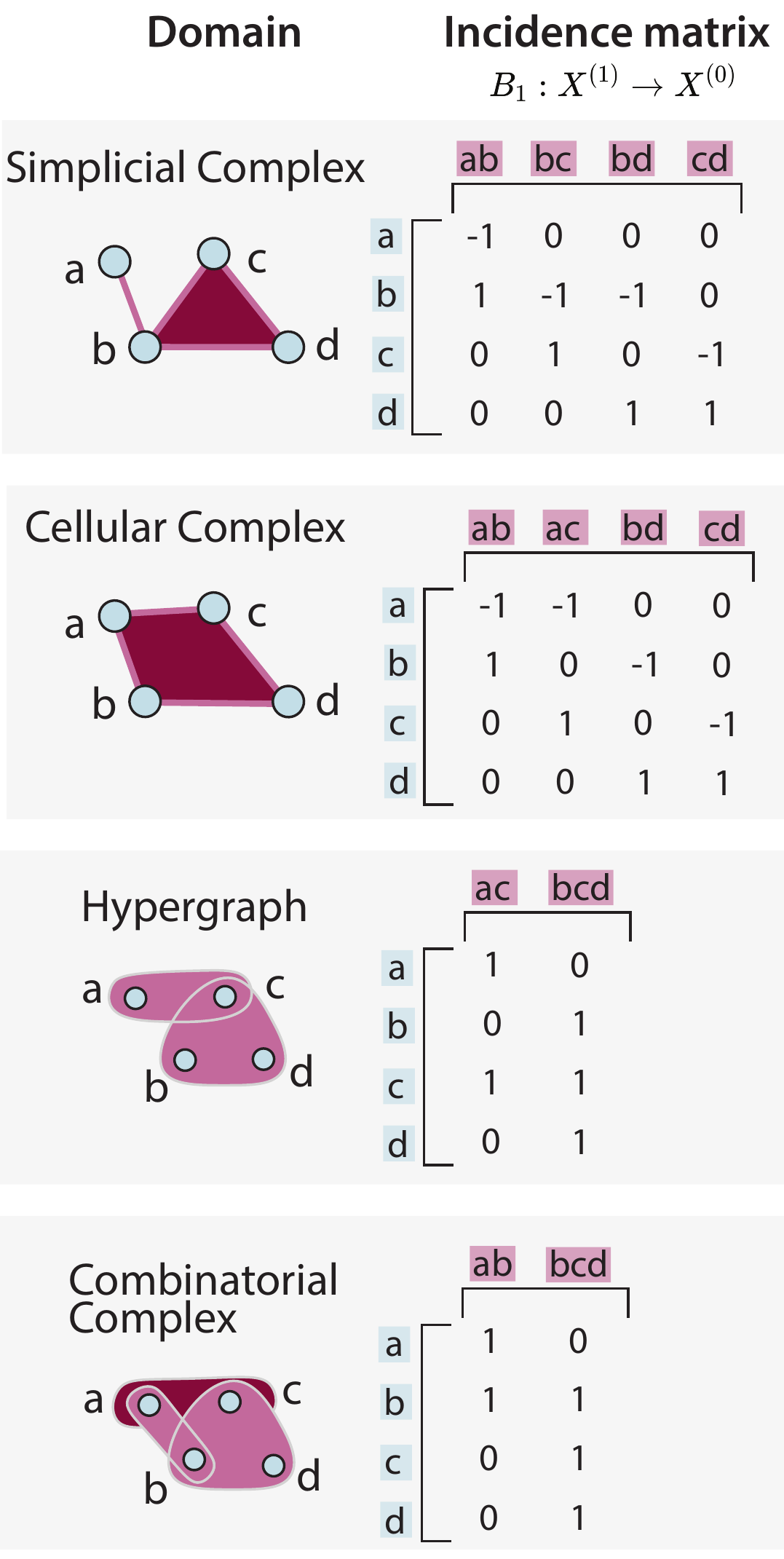}
	\caption{\textbf{Incidence Matrices.} Examples of an SC, a CC, an HG, and a CCC with their corresponding boundary matrices $B_1$ which map from 1-cells to 0-cells. The SC and CC maps are signed to encode edge orientation: the node appearing first in the arbitrary ordering (a,b,c,d) is always assigned -1.}
	\label{fig:boundaryex}
\end{figure}

A TNN successively updates cell features throughout its layers by using a notion of \textit{nearness} between cells: the \textit{neighborhood structure} (Figure \ref{fig:tnn}). Neighborhood structures are defined by \textbf{boundary relations}, which describe how cells of different ranks relate to each other. 
A cell $y$ of rank $r$ is said to be on the \textbf{boundary} of cell $x$ of rank $R$ if it is connected to $x$ and rank $r < R$. This relation is expressed as $y \prec x$. For example, a node connected to an edge is said to be on the boundary of that edge.

Boundary relations are encoded in \textbf{incidence matrices}. Specifically, we denote with $B_{r}$ the matrix that records which (regular) cells of rank $r-1$ bound which cells of rank $r$ (Figure \ref{fig:boundaryex}). Formally, $B_{r}$ is a matrix of size $n_{r-1} \cross n_{r}$, with $n_{r}$ denoting the number of cells of rank $r \geq 1$, defined: 

\begin{equation}
(B_r)_{i,j} = \begin{cases}
\pm 1 & x_i^{(r-1)} \prec x_j^{(r)} \\
0 & \text{otherwise,}
\end{cases}
\end{equation}

where $x_i^{(r-1)}$, $x_j^{(r)}$ are two cells of ranks $r-1$ and $r$ respectively. The $\pm 1$ sign encodes a notion of orientation required for SCs and CCs \cite{aschbacher1996combinatorial,klette2000cell}, and is always $+1$ for HGs and CCCs. 

\begin{figure*}[!ht]
    \centering
    \includegraphics[width=0.9\textwidth]{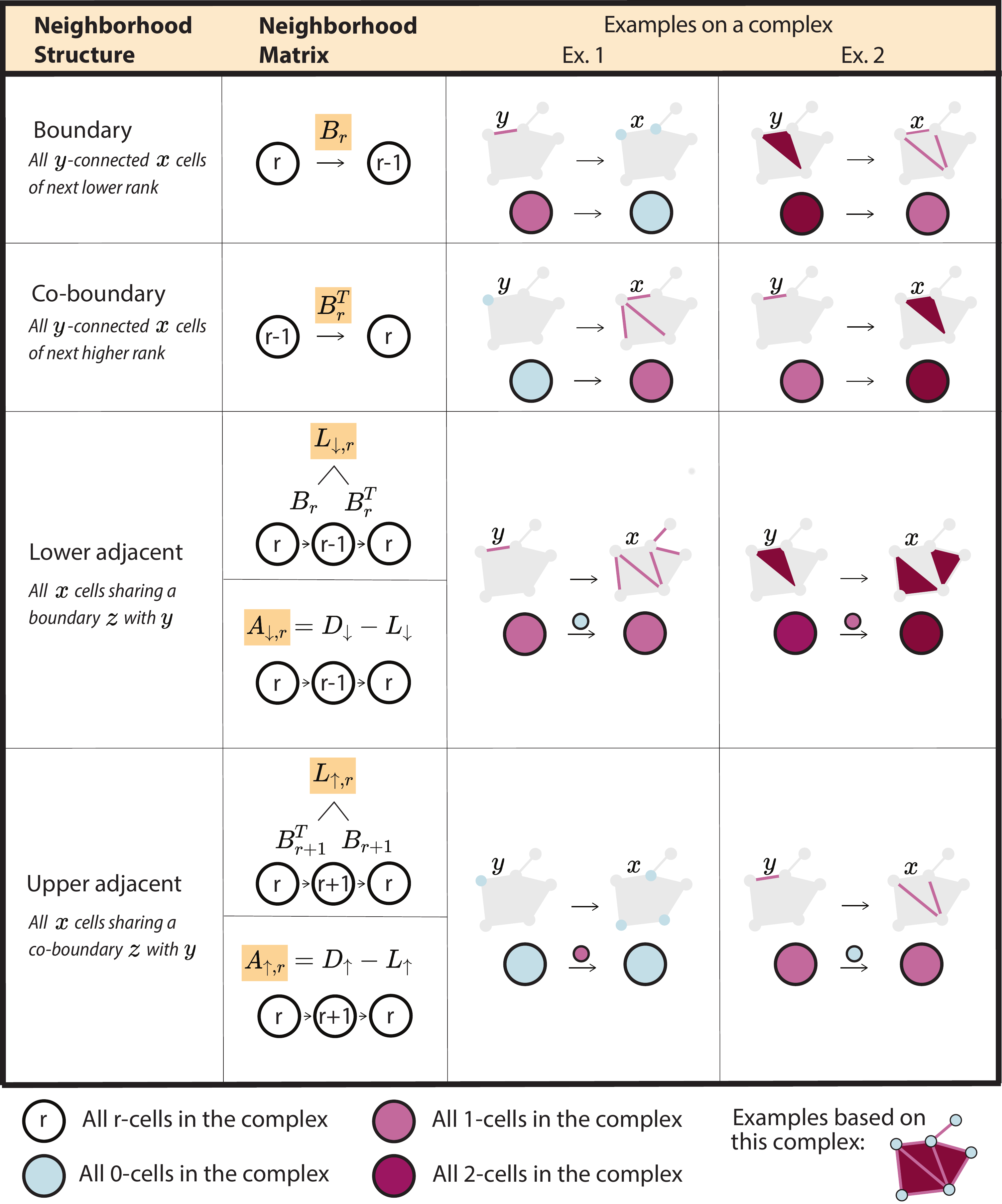}
	\caption{\textbf{Neighborhood Structures:} their neighborhood matrices and illustrations for a cell $x$ in the neighborhood of a cell $y$.}
	\label{fig:neighbors}
\end{figure*}

Incidence matrices can be used to encode the four most common neighborhood structures used in the literature, which we illustrate in Fig. \ref{fig:neighbors} and define below. Here, $L_{\uparrow,0}$ denotes the typical graph Laplacian. Its higher order generalization, the $r$-Hodge Laplacian, is $H_r = L_{\downarrow,r} + L_{\uparrow,r}$ \cite{barbarossasardellitti2020topological,schaub2021signal}. $D_r \in \mathbb{N}^{n_r \cross n_r}$ denotes the degree matrix, a diagonal matrix representing the number of connections of $r$-cells with $(r+1)$-cells.

\begin{tcolorbox}
    \begin{center}
        \normalsize \textbf{Neighborhood Structures}
    \end{center} \vspace{.25cm}

\small
\textbf{Boundary Adjacent Neighborhood} \\ \quad 
$\mathcal{B}(y)=\{x \mid x \prec y\}$: \\
The set of $y$-connected $x$ cells of next lower rank. The neighborhood is specified with the \textit{boundary matrix} $B_r$. \textit{Example: The set of nodes $x$ connected to edge $y$. }

\textbf{Co-Boundary Adjacent Neighborhood} \\ \quad $\mathcal{C}(y)=\{x \mid y \prec x\}$: \\
The set of $y$-connected $x$ cells of next higher rank. The neighborhood is specified with the \textit{co-boundary matrix} $B^T_r$. \textit{Example: The set of edges $x$ connected to node $y$.}

\textbf{Lower Adjacent Neighborhood}  \\ \quad $\mathcal{L}_{\downarrow}(y)=\{x \mid \exists z$ s.t. $z \prec y$ and $z \prec x\}$: \\
The set of $x$ cells that share a boundary $z$ with $y$. The neighborhood is specified with either the \textit{lower Laplacian matrix} $L_{\downarrow,r} = B_r B_r^T$ or the \textit{lower adjacency matrix} $A_{\downarrow,r} = D_r - L_{\downarrow,r}$. \textit{Example: the set of edges $x$ that connect to any of the nodes $z$ that touch edge $y$.} \\
\textbf{Upper Adjacent Neighborhood} \\  
\quad $\mathcal{L}_{\uparrow}(y)=\{x \mid \exists z$ s.t. $y \prec z$ and $x \prec z\}$:\\
The set of $x$ cells that share a co-boundary $z$ with $y$. The neighborhood is specified with either the \textit{upper Laplacian matrix} $L_{\uparrow,r} = B_{r+1}^T B_{r+1}$  or the \textit{upper adjacency matrix} $A_{\uparrow,r} = D_r - L_{\uparrow,r}$. \\ \textit{Example: The set of nodes $x$ that touch any of the edges $z$ that touch node $y$.}
\end{tcolorbox}

\subsection{Message Passing}
\textit{Message passing} defines the computation performed by a single layer $t$ of the TNN. During message passing, each cell's feature $\mathbf{h_x^{t, (r)}}$ is updated to incorporate: (1) the features associated with cells in its neighborhood and (2) the layer's learnable parameters denoted $\Theta^t$. The term ``message passing'' reflects that a signal is ``traveling'' through the network, passing between cells on paths laid out by the neighborhood structure. The output $\mathbf{h}^{t+1}$ of layer $t$ becomes the input to layer $t+1$. In this way, deeper layers incorporate information from more distant cells, as information diffuses through the network.
\subsubsection{The Steps of Message Passing}
\label{sec:message-passing}
\begin{figure}
	\centering
 	\includegraphics[width=0.9\linewidth ]{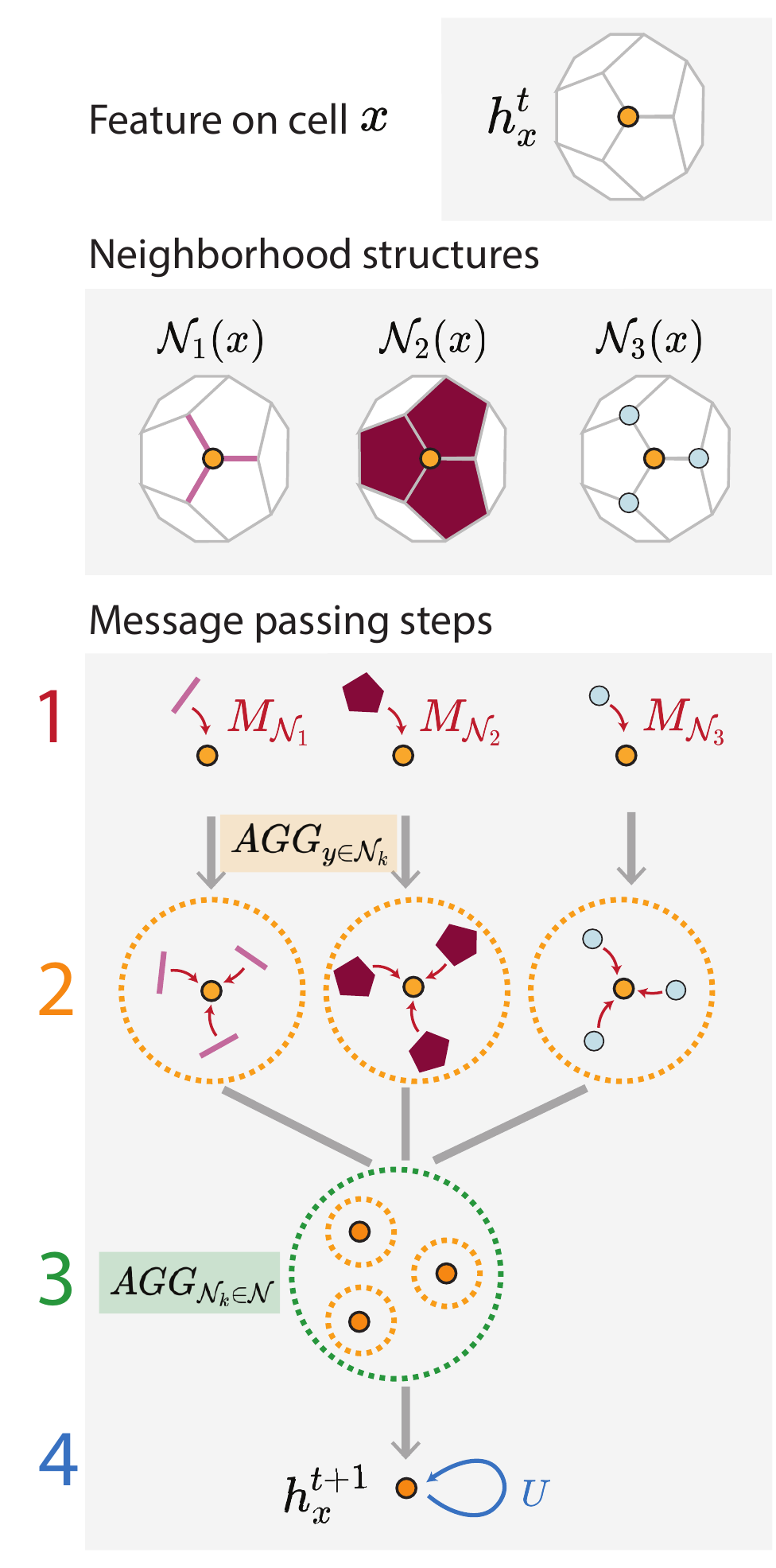}
	\caption{\textbf{Message passing steps:} 1: Message (red), 2: Within-neighborhood aggregation (orange), 3: Between-neighborhood aggregation (green), 4: Update (blue). The scheme updates a feature $\mathbf{h}_x^{t,(r)}$ on a $r$-cell $x$ at layer $t$ (left column) into a new feature $\mathbf{h}_x^{t+1,(r)}$ on that same cell at the next layer $t+1$ (right column). Here, the scheme uses four neighborhood structures $\mathcal{N}_k$ for $k \in \{1, 2, 3, 4\}$ (middle column). Figure adapted from \cite{hajij2023tdl}. \vspace{-.5cm}}
	\label{fig:messageoverview}
\end{figure}

We decompose message passing into four steps, adopted from the framework of \cite{hajij2023tdl}. Each step is represented with a different color\textemdash \textit{red}, \textit{orange}, \textit{green}, or \textit{blue}\textemdash illustrated in Figure  \ref{fig:messageoverview}.

\clearpage
\clearpage

\begin{tcolorbox}
\begin{center}
    \textbf{The Steps of Message Passing}
\end{center} \vspace{.25cm}

\small 
\noindent \textbf{1. Message (red):} First, a message $m_{y \rightarrow x}^{\left(r^{\prime} \rightarrow r\right)}$ travels from a $r'$-cell $y$ to a $r$-cell $x$ through a neighborhood $k$ of $x$ denoted $\mathcal{N}_k (x)$:
\vspace{-1mm}
\begin{align}
 {m_{y \rightarrow x}^{\left(r^{\prime} \rightarrow r\right)}}&=M_{\mathcal{N}_k}\left(\mathbf{h}_x^{t,(r)}, \mathbf{h}_y^{t,(r^{\prime})}, \Theta^t \right).
\end{align}
via the function $M_{\mathcal{N}_k}$ depicted in red in Figure \ref{fig:messageoverview}. Here, $\textbf{h}_x^{t,(r)}$ and $\textbf{h}_y^{t,(r')}$ are features of dimension $d_r$ and $d_{r'}$ on cells $y$ and $x$ respectively, and $\Theta^t$ are learnable parameters. In the simplest case, this step looks like a neighborhood matrix $M$ propagating a feature $\mathbf{h}_y^{t,(r')}$ on $r'$-cell $y$ to $r$-cell $x$ as: 
\begin{align}
    m_{y \rightarrow x}^{\left(r^{\prime} \rightarrow r\right)}&= M_{xy}\cdot \textbf{h}_y^{t,(r')} \cdot \Theta^t,
\end{align}
where $M_{xy}$ is the scalar entry of matrix $M$ at the row corresponding to cell $x$ and column corresponding to cell $y$ and $m_{y \rightarrow x}^{\left(r^{\prime} \rightarrow r\right)}$ and $\Theta$ is a $d_{r'} \times d_{r}$ matrix. If $y$ is not in the neighborhood structure of $x$, then $M_{xy}$ will be 0, and $x$ cannot receive any message from $y$. \\

\noindent \textbf{2. Within-Neighborhood Aggregation (orange):}
Next, messages are aggregated across all cells $y$ belonging to the neighborhood $\mathcal{N}_k(x)$:\vspace{-1mm}
\begin{align}
 {m_x^{\left(r^{\prime} \rightarrow r\right)}}&=A G G_{y \in \mathcal{N}_k(x)} m_{y \rightarrow x}^{\left(r^{\prime} \rightarrow r\right)}, 
 \end{align}
resulting in the \textit{within-neighborhood aggregated message} $m_x^{\left(r^{\prime} \rightarrow r\right)}$. Here, $AGG$ is an aggregation function, depicted in orange in Figure \ref{fig:messageoverview}, analogous to pooling in standard convolutional networks. \\

\noindent \textbf{3. Between-Neighborhood Aggregation (green):}
Then, messages are aggregated across neighborhoods in a neighborhood set $\mathcal{N}$: \vspace{-1mm}
 \begin{align}
{m_x^{(r)}}&=A G G_{\mathcal{N}_k \in \mathcal{N}} m_x^{\left(r^{\prime} \rightarrow r\right)},
\end{align}
where AGG is a (potentially different) aggregation function depicted in green in Figure  \ref{fig:messageoverview}, and $m_x^{(r)}$ is the message received by cell $x$ that triggers the update of its feature. \\ 

\noindent \textbf{4. Update (blue):}
Finally, the feature on cell $x$ is updated via a function $U$ depicted in blue in Figure  \ref{fig:messageoverview}, which may depend on the previous feature $\textbf{h}_x^{t,(r)}$ on cell $x$:\vspace{-1mm}
\begin{align}
{\textbf{h}_x^{t+1,(r)}}&=U\left(\textbf{h}_x^{t,(r)}, m_x^{(r)}\right),
\end{align}
The result $\textbf{h}_x^{t+1,(r)}$ is the updated feature on cell $x$ that is input to layer $t+1$.
\end{tcolorbox}

\vspace{.5cm}

In this review, we decompose the structure of TNN architectures proposed in the literature into these four message passing steps\textemdash a unified notational framework adopted from \cite{hajij2023tdl} that allows us to contrast existing approaches. Many architectures repeat steps and/or modify their order.
We note that this conceptualization of message passing as a local, cell-specific operation is called the \textit{spatial approach} \cite{gilmer2017neural}. In GNNs and TNNs alike, message passing can alternatively be expressed in its dual \textit{spectral} form, using global Fourier analysis over the domain. For this review, we choose to write all equations in spatial form for intuitiveness and generality \cite{bodnar2021topological,hajij2020cell,heydari2022message}.

\subsubsection{Tensor Diagrams}

\begin{figure*}[!ht]
	\centering
 	\includegraphics[width=1.\textwidth ]{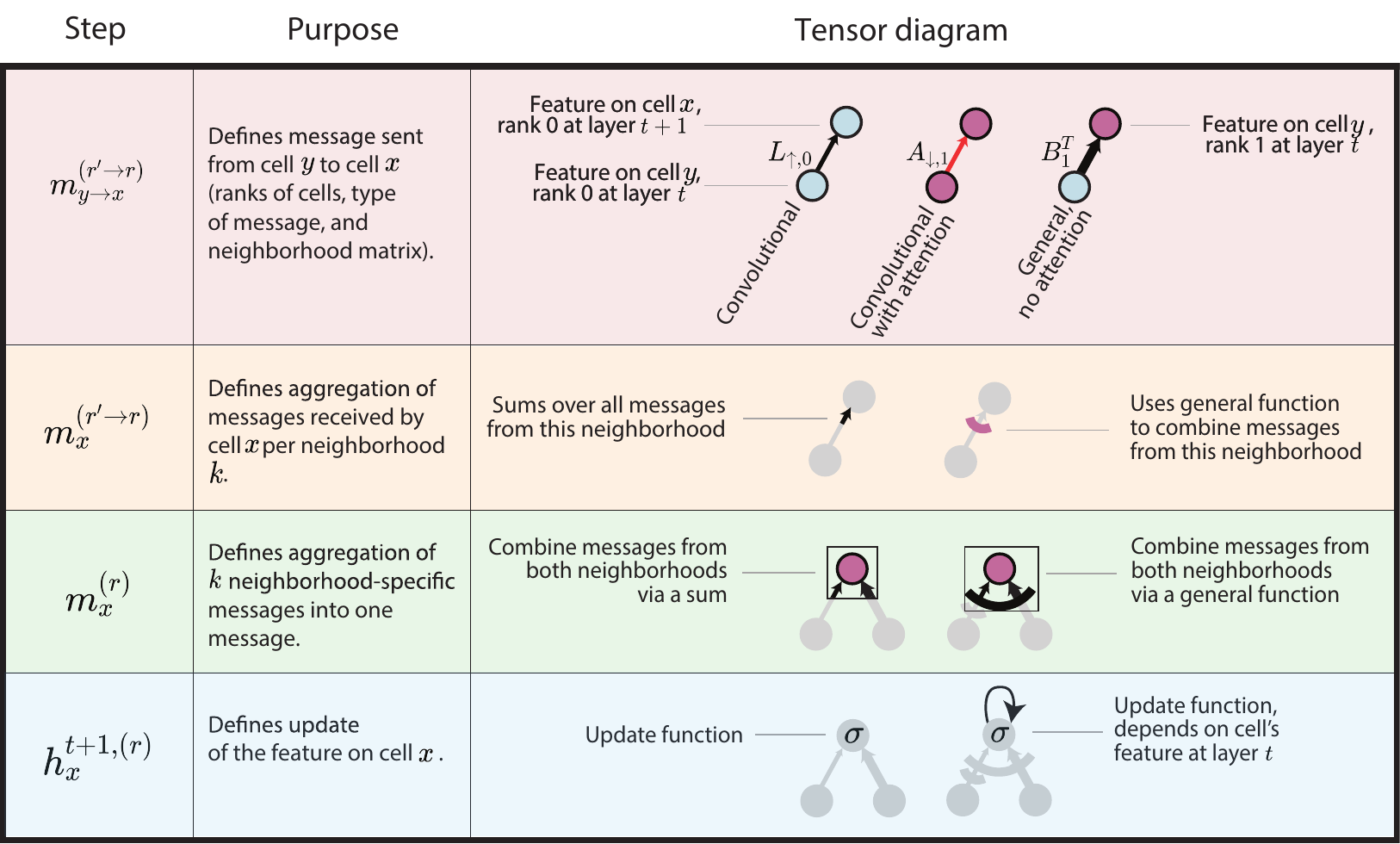}
	\caption{\textbf{Tensor Diagrams:} a graphical notation for the four steps of a message passing scheme. A diagram depicts how a feature on cell $y$ at layer $t$, $\textbf{h}_y^{(t)}$, becomes a feature on cell $x$ at layer $t+1$, $\textbf{h}_x^{(t+1)}$.}
	\label{fig:messagesteps}
\end{figure*}

We visually represent message passing schemes with an adapted version of the \textit{tensor diagram} introduced in \cite{hajij2022higher} and further developed in \cite{hajij2023tdl}. A tensor diagram provides a graphical representation of a TNN architecture. Figure \ref{fig:messagesteps} explains the recipe for constructing a tensor diagram from message passing steps. 

\subsubsection{Types of Message Passing Functions}

\begin{figure*}[!ht]
	\centering
 	\includegraphics[width=0.7\textwidth ]{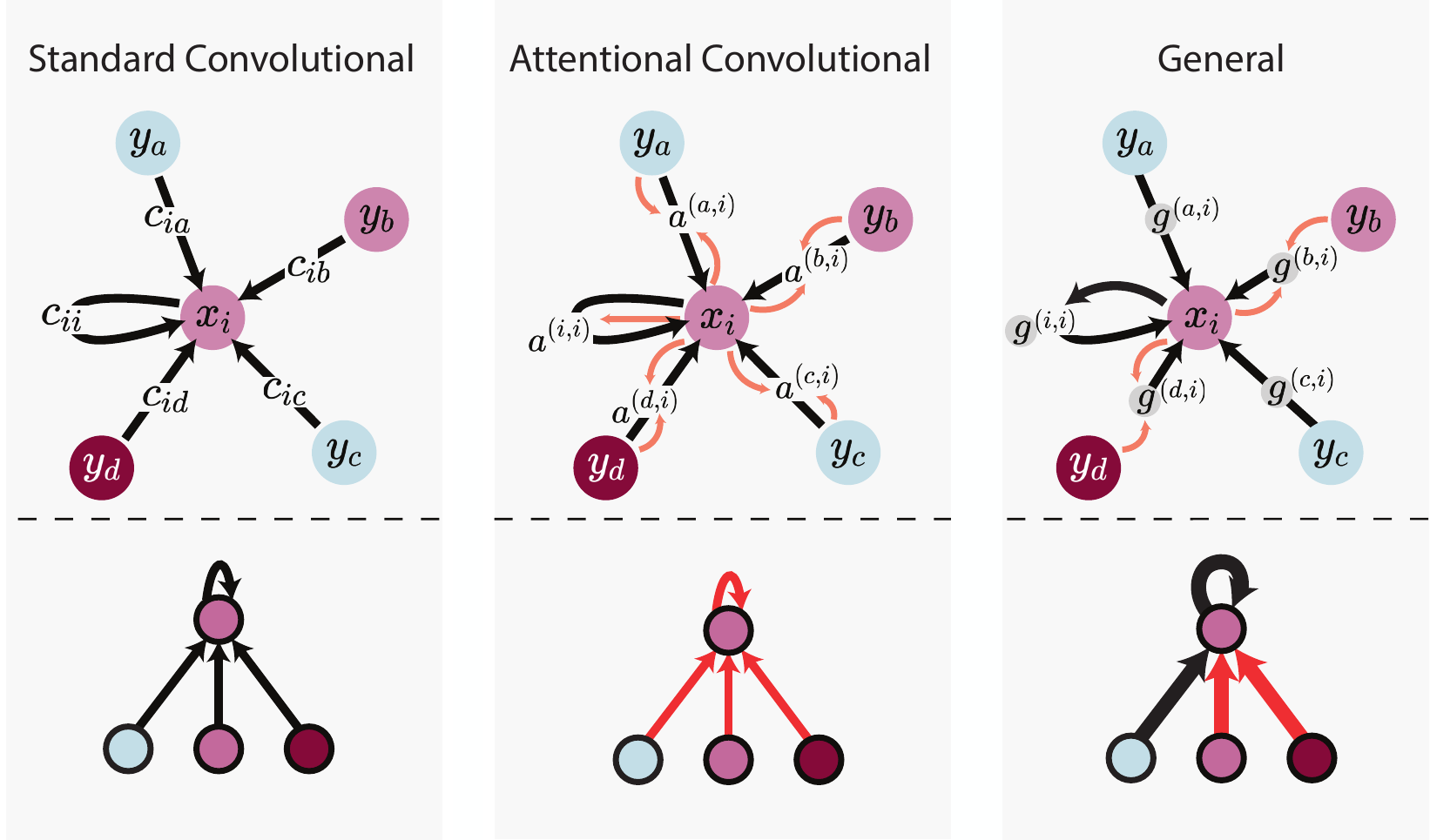}
	\caption{\textbf{Types of Message Passing Functions.} In each case, a cell $x_i$ (an edge) receives information from its various neighbors, cells $y_j$ (two nodes, an edge, and a face). The message received by cell $x_i$ from cell $y_j$ is determined by a specific function $c(x_i,y_j)$, $a(x_i,y_j)$, or $g(x_i,y_j)$. Top: Each neighborhood cell $y_j$ sends a message to cell $x_i$. (Inspired by P. Veličković and \cite{bronsteinblog}). Bottom: Illustration of the message-passing scheme above using tensor diagrams \cite{hajij2023tdl}. \vspace{-.5cm}}
	\label{fig:messagetypes}
\end{figure*}

The message passing function $M_{\mathcal{N}_k}$ employed in Step 1 is defined by the practitioner. There are three kinds of functions commonly used in the literature, as outlined in Figure \ref{fig:messagetypes} \cite{bronsteinblog}. The variety used determines how layer parameters weight each incoming message from cell $y$ to cell $x$. The \textit{standard convolutional} case multiplies each message by some learned scalar. The \textit{attentional convolutional} case weights this multiplication depending on the features of the cells involved. The \textit{general} case implements a potentially non-linear function that may or may not incorporate attention. Some schemes also make use of fixed, non-learned weights to assign different levels of importance to higher-order cells. Figure \ref{fig:messagetypes} illustrates each type with tensor diagrams.

\section{Literature Review}\label{sec4}
 We now review the literature on topological neural networks (TNNs) over hypergraphs, simplicial complexes, cellular complexes, and combinatorial complexes, using the conceptual framework of Section \ref{sec3}. We summarize and compare the TNNs in terms of their architectures (Section \ref{sec:architectures}), the machine learning tasks to which they have been applied (Section \ref{sec:applications}), and their geometric properties (Section \ref{sec:symmetries}).
 
 \subsection{Architectures}
 \label{sec:architectures}
\begin{figure*}
    \centering
    \includegraphics[width=0.7 \textwidth]{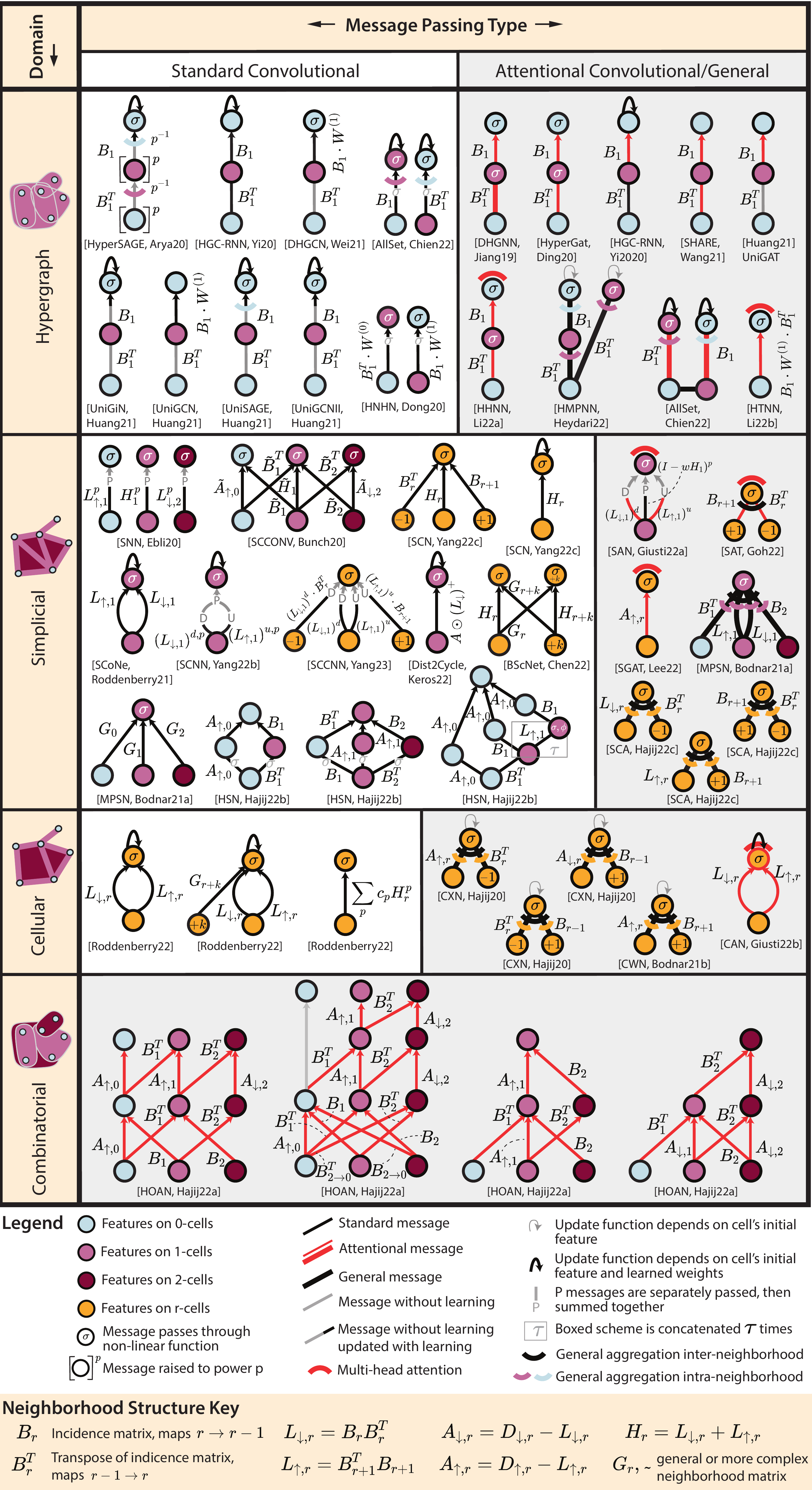}
    \caption{\textbf{Topological Neural Networks (TNNs): A Graphical Literature Review.} We organize TNNs according to the \textit{domain} (rows) and the \textit{message passing type} (columns).}
    \label{fig:lit_table}
\end{figure*}

Figure \ref{fig:lit_table} summarizes TNN architectures according to the fundamental concepts introduced in Section \ref{sec3}, with the \textit{domain} on the vertical axis, the \textit{message passing type} on the horizontal axis, \textit{neighborhood structures} and \textit{message passing equations} visually represented with tensor diagrams. We share complete message passing equations for each architecture\textemdash decomposed according to the four steps introduced in Section \ref{sec:message-passing} and rewritten in unifying notations \textemdash at \url{github.com/awesome-tnns}.

\subsubsection{Hypergraphs}

Of the domains considered here, hypergraph neural networks have been most extensively researched, and have been surveyed previously \cite{ling2021review,gao2022hypergraph,hu2021survey,wang2021session,fischer2021towards}. Many papers in the early literature do not use hypergraphs as the computational domain. Rather, algorithms like clique-expansion \cite{zien1999multilevel,agarwal2005beyond,zhou2006learning} are used to reduce hypergraphs to graphs, which are then processed by the model. This reduction adversely affects performance, as structural information is lost \cite{hein2013total,li2013z,chien2019hs}. Many such graph-based models\textemdash including HGNN \cite{feng2019hypergraph}, HyperConv\cite{bai2021hypergraph}, HyperGCN \cite{yadati2019hypergcn}, and HNHN \cite{dong2020hnhn}\textemdash are used as benchmarks for more recent models that \textit{do} computationally operate on hypergraphs. Here, we focus on models that preserve hypergraph structure during learning.

Many hypergraph models use a message passing scheme comprised of two phases, with information flowing from nodes to their hyperedges and then back to the nodes. We call this the \textit{two-phase scheme}. The scheme appears in many tensor diagrams of Figure \ref{fig:lit_table} where information flows from blue to pink (phase 1) and then from pink to blue (phase 2). The scheme is used in models with both standard and attentional message passing.

\textbf{Standard}

Of those using a standard message passing, the models from \cite{arya2020hypersage}, \cite{yi2020hypergraph}, \cite{wei2021dynamic}, and \cite{huang2021unignn} use the two-phase scheme. \cite{yi2020hypergraph} is unique in using a learnable weight matrix in the first phase of message passing. On the second phase, \cite{wei2021dynamic} and the UniGCN model from \cite{huang2021unignn} are unique in using a fixed weight matrix on top of learnable weights. In \cite{arya2019hyperlearn}, \cite{yi2020hypergraph}, and the UniGNN, UniSAGE, and UniGCNII models from \cite{huang2021unignn}, the initial feature on each node is recurrently used to update each incoming message\textemdash denoted with a looped black arrow in Figure \ref{fig:lit_table}. We note that \cite{huang2021unignn} systematically generalizes some of the most popular GNN architectures to hypergraphs with its unifying framework: UniGNN.

In \cite{dong2020hnhn}, fixed weights are used on both the node to hyperedge and hyperedge to node phases. The paper AllSet \cite{chien2022you} uses a similar structure while incorporating fully learnable multi-set functions for neighborhood aggregation, which imbues its TNNs with high expressivity and generality. EHNN \cite{kim2022equivariant} (excluded from Figure \ref{fig:lit_table} for its complexity; see written equations) proposes a maximally expressive model using sparse symmetric tensors to process data on hypergraphs with uniformly sized hyperedges. 

\textbf{Attentional / General}

The models from \cite{jiang2019dynamic}, \cite{ding2020less}, \cite{yi2020hypergraph}, \cite{wang2021session}, and the UniGAT model from \cite{huang2021unignn} employ the two-phase scheme in concert with attentional message passing. The architectures of \cite{li2022heterogeneous} and \cite{li2022hypergraph} apply multi-head attention. \cite{heydari2022message} adapts the two-phase scheme in order to update node features through two parallel paths. \cite{chien2022you} and \cite{kim2022equivariant} offer transformer-based variants of their standard architectures, concurrently with \cite{li2022hypergraph}.


\subsubsection{Simplicial Complexes}

Simplicial complexes were first explored from a signal processing perspective \cite{battiston2020networks}, with initial focus on edge flows \cite{jiang2011statistical,schaub2018flow}, Hodge Laplacians \cite{barbarossasardellitti2020topological,schaub2020random}, and convolution \cite{yang2021finite,yang2022simplicial,isufi2022convolutional}. As a precursor to deep learning, \cite{roddenberry2019hodgenet} introduced $\mathcal{L}_{\downarrow,1}$ in HodgeNet to learn convolutions on edge features on graphs. This contrasts with former GNN approaches processing node features.

\textbf{Standard}

\cite{ebli2020simplicial} (SNN) and \cite{bunch2020simplicial} (SCCONV) first generalized the convolutional approach of \cite{roddenberry2019hodgenet} to features supported on faces and cells of higher ranks. Unlike HodgeNet, SNN and SCCONV use both $\mathcal{L}_{\downarrow,1}$ and $\mathcal{L}_{\uparrow,1}$. In SNN \cite{ebli2020simplicial} messages are not passed between adjacent ranks. By contrast, SCCONV uses independent boundary and co-boundary neighborhoods, hence incorporating features from adjacent ranks. \cite{yang2022efficient} also makes use of this multi-neighborhood scheme for updating and classifying edge features. They also propose a single neighborhood scheme with an update including the initial cell's feature. \cite{roddenberryglaze2021principled} and \cite{yang2022simplicialnn} devise schemes where messages coming from $\mathcal{L}_{\downarrow,1}$ and $\mathcal{L}_{\uparrow,1}$ are weighted separately, providing greater learning flexibility. \cite{yang2022simplicialnn} allows features to travel multiple hops through the domain by using a polynomial form of the neighborhood structures, leveraging the simplicial convolutional filter from\cite{yang2022simplicial}. \cite{yang2023convolutional} extend this multiple-hop model with additional neighborhood structures. \cite{keros2022dist2cycle} used a modified version of $\mathcal{L}_\downarrow$ to find signals coiled around holes in the complex. BSCNet \cite{chen2022bscnets} combines node and edge-level shifting to predict links between nodes. This was the first model to pass messages between arbitrary ranks, leveraging a pseudo Hodge Laplacian.

MPSN \cite{bodnar2021topological} explicitly details their message-passing scheme in the spatial domain, subsuming previous models described from a spectral approach. \cite{hajij2022high} introduces High Skip Networks (HSNs), in which each layer updates features through multiple sequential higher-order message passing steps, ``skipping'' it through higher ranks as a generalization of skip-connections in conventional neural networks.

\textbf{Attentional /  General}

SAN \cite{giusti2022simplicial}, SAT \cite{goh2022simplicial}, and SGAT \cite{lee2022sgatsg} concurrently introduced attentional message passing networks on the simplicial domain. Each model makes use of a unique set of neighborhood structures and attention coefficients. SGAT is the only model as of yet developed for heterogeneous simplicial complexes of general rank. \cite{hajij2022simplicial} introduces a variety of general message passing schemes with two neighborhood structures. \cite{bodnar2021topological} uses all four neighborhood structures, endowing each with a separate learnable matrix and general aggregation function.

\subsubsection{Cellular Complexes}

Just as for simplicial complexes, cellular complex networks have been significantly influenced by work in signal processing \cite{barbarossasardellitti2020topological, sardellitti2021topological,roddenberry2022signal}. These works demonstrated that representing data in the CC domain yields substantially better results than the more rigid SC domain.

\textbf{Standard}

\cite{roddenberry2022signal} proposes theoretically possible message passing schemes for CCs inspired by works in the SC domain. As of yet, these models have not been implemented.

\textbf{Attentional / General}

\cite{hajij2020cell} introduces the first TNNs to be theoretically defined on the CC domain. \cite{bodnar2021weisfeiler} was the first to implement and evaluate such a model, and demonstrated that TNNs on CCs outperform state-of-the-art graph-based models in expressivity and classification tests. The CAN model from \cite{giusti2022cell} adapts a modified version of the message passing scheme from \cite{giusti2022simplicial} onto the CC domain.

\subsubsection{Combinatorial Complexes}
The combinatorial complex domain was only recently mathematically defined by \cite{hajij2022higher}. This work introduces four attentional message passing schemes for CCCs tailored to mesh and graph classification. A more extensive analysis is needed to quantify the advantages of this domain over other topological domains.

\subsection{Tasks}
\label{sec:applications}
Table \ref{tab:applications} reviews the tasks studied by each paper proposing TNNs. Tasks are first categorized into: \textit{node-level} tasks assigning labels to nodes, as in node classification, regression or clustering; \textit{edge-level tasks} assigning labels to edges, as in edge classification or link prediction; and \textit{complex-level} tasks assigning labels to each complex as a whole, as in hypergraph classification. Tasks are additionally labeled according to their purpose (e.g. classification, regression, prediction). We also indicate the extent of benchmarking performed on each model and code availability.

\subsection{Symmetries and Geometric Properties}\label{sec:symmetries}

Topological domains possess symmetries and other geometric properties that should be respected to ensure the quality of the features learned by a TNN \cite{bronstein2017geometric}. Here, we outline such properties harnessed by models in the literature.

\paragraph{Hypergraphs.} On hypergraphs, the following symmetries are desirable:
\begin{enumerate}
    \item \textit{Permutation Invariance:} Relabeling the nodes and applying the TNN yields an output that is identical to the original output obtained without relabeling. This requires the aggregation functions to be permutation invariant, such as a mean or a sum \cite{arya2020hypersage, kim2022equivariant, chien2022you, dong2020hnhn, keros2022dist2cycle}. This is also called \textit{hypergraph isomorphism invariance}.
\end{enumerate}
\newgeometry{left=3cm,right=3cm}
{
\setcounter{table}{0}

\begin{table}
\centering
\footnotesize
\begin{minipage}{0.9\textwidth}
\caption{
\textbf{Applications of Topological Neural Networks (TNNs)}. We organize papers according to domain and task level, task purpose, and extent of benchmark testing (Graph: compared to graph-based models, GNN SOTA: compared to GNN state-of-the-art, TNN SOTA: compared to state-of-the-art on topological domain). We exclude papers without implementation, and use * to indicate that an implementation has not been shared.\label{tab:applications}
}\end{minipage}
\begin{tabular}{lllllll}

\textbf{Domain} &
  \textbf{Model} &
  \multicolumn{3}{c}{\textbf{Task Level}} &
  \textbf{Task Purpose} &
  \textbf{Comparisons} \\ \midrule

 &
   & \rule{0pt}{33pt} 
  \begin{rotate}{90}\textit{Node} \end{rotate}&
  \begin{rotate}{90}\textit{Edge} \end{rotate} &
  \begin{rotate}{90}\textit{Complex} \end{rotate}&
   &
   \\ \toprule
\textbf{HG}&
  HyperSage \cite{arya2020hypersage} &
  \checkmark &
   &
   &
  \begin{tabular}[c]{@{}l@{}}Classification (Inductive + Transductive)\end{tabular} &
  GNN SOTA \\
  \cmidrule(lr){2-7}
 &
  AllSet \cite{chien2022you} &
  \checkmark &
   &
   &
  Classification &
  TNN SOTA \\
  \cmidrule(lr){2-7}

 &
  HyperGat \cite{ding2020less} &
  \checkmark &
   &
   &
  Classification &
  GNN SOTA \\
  \cmidrule(lr){2-7}

 &
  HNHN \cite{dong2020hnhn} &
  \checkmark &
  \checkmark &
   &
  \begin{tabular}[c]{@{}l@{}}Classification, Dimensionality Reduction\end{tabular} &
  GNN SOTA \\
    \cmidrule(lr){2-7}

 &
  HMPNN* \cite{heydari2022message} &
  \checkmark &
   &
   &
  Classification &
  TNN SOTA \\
    \cmidrule(lr){2-7}

 &
  UniGNN \cite{huang2021unignn} &
  \checkmark &
   &
   &
  \begin{tabular}[c]{@{}l@{}}Classification (Inductive + Transductive)\end{tabular} &
  TNN SOTA \\
    \cmidrule(lr){2-7}

 &
  DHGNN \cite{jiang2019dynamic} &
  \checkmark &
   &
   &
  \begin{tabular}[c]{@{}l@{}}Classification (Multimodal)\end{tabular} &
  GNN SOTA \\
    \cmidrule(lr){2-7}

 &
  EHNN \cite{kim2022equivariant} &
  \checkmark &
   &
   &
  \begin{tabular}[c]{@{}l@{}}Classification, Keypoint Matching\end{tabular} &
  TNN SOTA \\
    \cmidrule(lr){2-7}

 &
  HHNN \cite{li2022heterogeneous} &
  \checkmark &
   &
   &
  Link prediction &
  TNN SOTA \\
    \cmidrule(lr){2-7}

 &
  HTNN \cite{li2022hypergraph} &
  \checkmark &
   &
   &
  Classification &
  TNN SOTA \\
    \cmidrule(lr){2-7}

 &
  SHARE* \cite{wang2021session} &
  \checkmark &
   &
   &
  Prediction &
  GNN SOTA \\
    \cmidrule(lr){2-7}

 &
  DHGCN* \cite{wei2021dynamic} &
   &
   &
  \checkmark &
  Classification &
  GNN SOTA \\
    \cmidrule(lr){2-7}

 &
  HGC-RNN* \cite{yi2020hypergraph} &
  \checkmark &
   &
   &
  Prediction &
  GNN SOTA \\ \toprule
\textbf{SC} &
  MPSN \cite{bodnar2021topological} &
   &
  \checkmark &
  \checkmark &
  \begin{tabular}[c]{@{}l@{}}Classification, Trajectory Classification\end{tabular} &
  GNN SOTA \\
    \cmidrule(lr){2-7}

 &
  SCCONV \cite{bunch2020simplicial} &
   &
   &
  \checkmark &
  Classification &
  Graph \\
    \cmidrule(lr){2-7}

 &
  BScNet \cite{chen2022bscnets} &
   &
  \checkmark &
   &
  Link prediction &
  GNN SOTA \\
    \cmidrule(lr){2-7}

 &
  SNN \cite{ebli2020simplicial} &
   &
  \checkmark &
   &
  Imputation & None
   \\
     \cmidrule(lr){2-7}

 &
  SAN \cite{giusti2022simplicial} &
   &
  \checkmark &
   &
  \begin{tabular}[c]{@{}l@{}}Classification, Trajectory Classification\end{tabular} &
  TNN SOTA \\
    \cmidrule(lr){2-7}

 &
  SAT \cite{goh2022simplicial} &
   &
  \checkmark &
  \checkmark &
  \begin{tabular}[c]{@{}l@{}}Classification, Trajectory Classification\end{tabular} &
  TNN SOTA \\
    \cmidrule(lr){2-7}

 &
  HSN* \cite{hajij2022high} &
  \checkmark &
  \checkmark &
  \checkmark &
  \begin{tabular}[c]{@{}l@{}}Classification, Link prediction, Vector embedding\end{tabular} &
  Graph \\
    \cmidrule(lr){2-7}

 &
  SCA* \cite{hajij2022simplicial} &
   &
   &
  \checkmark &
  Clustering &
  Graph \\
    \cmidrule(lr){2-7}

 &
  Dist2Cycle \cite{keros2022dist2cycle} &
   &
  \checkmark &
   &
  Homology Localization &
  GNN SOTA \\
    \cmidrule(lr){2-7}

 &
  SGAT \cite{lee2022sgatsg} &
  \checkmark &
   &
   &
  Classification &
  GNN SOTA \\
    \cmidrule(lr){2-7}

 &
  SCoNe \cite{roddenberryglaze2021principled} &
   &
  \checkmark &
   &
  Trajectory Classification &
  TNN SOTA \\
    \cmidrule(lr){2-7}

 &
  SCNN* \cite{yang2022simplicial} &
   &
  \checkmark &
   &
  Imputation &
  TNN SOTA \\
    \cmidrule(lr){2-7}

 &
  SCCNN \cite{yang2023convolutional} &
   &
  \checkmark &
   &
  \begin{tabular}[c]{@{}l@{}}Link prediction, Trajectory Classification\end{tabular} &
  TNN SOTA \\
    \cmidrule(lr){2-7}

 &
  SCN \cite{yang2022efficient} &
   &
  \checkmark &
   &
  Classification &
  TNN SOTA \\ \toprule
\textbf{CC}&
  CWN \cite{bodnar2021weisfeiler} &
   &
  \checkmark &
  \checkmark &
  \begin{tabular}[c]{@{}l@{}}Classification, prediction, regression\end{tabular} &
  GNN SOTA \\
    \cmidrule(lr){2-7}
 &
  CAN \cite{giusti2022cell} &
   &
   &
  \checkmark &
  Classification &
  GNN SOTA \\ \toprule
\textbf{CCC}&
  HOAN* \cite{hajij2022higher} &
   &
   \checkmark& 
  \checkmark &
  Classification &
GNN SOTA \\ \bottomrule \\
  \vspace{0.5cm}

\end{tabular}
\end{table}
}
\restoregeometry
\begin{enumerate}
    \setcounter{enumi}{1}
    \item \textit{Global Neighborhood Invariance:} The network's representation of a node is invariant to hyperedge cardinality: a hyperedge connecting many nodes is weighted the same as a hyperedge connecting less nodes \cite{arya2020hypersage}.
\end{enumerate}

\paragraph{Simplicial Complex.} 
For simplicial complexes, the following symmetries have been considered:
\begin{enumerate}
    \item \textit{Permutation Invariance:} Invariance to node relabeling; the same as for HGs. \cite{schaub2021signal,roddenberryglaze2021principled,bodnar2021topological}
    \item \textit{Orientation Equivariance:} Changing the orientation of the simplicial complex (i.e. flipping the signs in the incidence matrix) re-orients the output of that network accordingly \cite{schaub2021signal,roddenberryglaze2021principled,bodnar2021topological}.
    \item \textit{Simplicial Locality (geometric property):} In each layer, messages are only passed between $r$-cells and $(r\pm 1)$-cells \cite{schaub2021signal}. If that property is not verified, and messages can pass between any $r$- and $r'$-cells, then the network has \textit{extended simplicial locality}.
\end{enumerate}

In addition, \textit{simplicial awareness} can be imposed, such that message passing on a simplicial complex with maximum cell rank $r$ depends on every rank $r' \leq r$ \cite{roddenberryglaze2021principled}. 

\paragraph{Cellular Complex and Combinatorial Complex.} Permutation invariance is defined for CCs \cite{bodnar2021weisfeiler} and CCCs \cite{hajij2022higher} just as for SCs and HGs. Beyond generalizing global neighborhood invariance to CCC, more research is required to understand the symmetries that can equip this general topological domain.

\section{Discussion}\label{discussion}
Our literature review has revealed the diversity of TNN architectures as well as their main axes of comparison. Looking to the future, we highlight four salient opportunities for development.

\paragraph{Within-Domain and Between-Domain Benchmarking.} 
Table \ref{tab:applications} shows that the domain choice strongly correlates with a TNN's task level. This necessarily makes within-domain comparisons difficult, regardless of code sharing. We also emphasize that many TNNs are only benchmarked against graph-based models or early models in their respective domain, which makes between-domain comparisons equally difficult. As the field grows, improving within and between-domain benchmarking mechanisms will be critical to better informing model selection and quantifying progress.

\paragraph{TNN Architectures on General Domains.} 

 The diversity of implementations on HGs and SCs point to a strong potential for similar development in the cellular and combinatorial domains. For instance, only one attentional CC model has been proposed \cite{giusti2022cell}. Moreover, any previously developed HG/SC/CC model can be reproduced in the CCC domain and, if desirable, improved with greater flexibility. Evaluating the impact of this added flexibility will directly characterize utility of richer topological structure in deep learning.

\paragraph{Connecting to the Graph Literature.} 

The HG field's ties to the graph community has led to GNN-based advancements not yet propagated to other domains. A first example are dynamic domains, successful with HGs for tasks like pose estimation \cite{liu2020semi}, rail transit modeling \cite{wang2021metro}, and co-authorship prediction \cite{jiang2019dynamic}. No work in other discrete domains has explored dynamism. In addition, outside of the HG domain, TNNs are largely implemented as homogeneous networks. This leaves room for heterogeneous and non-Euclidean generalizations. 

\paragraph{Going Deeper.} 

Over-smoothing occurs when a network is too effective at aggregating signal over multiple layers. This leads to very similar features across cells and poor performance on the downstream learning task. While this issue draws attention in the graph community \cite{chen2020measuring, oono2020graph,rusch2023survey}, little of this work has been generalized to TNNs, causing them to remain mostly shallow. UniGCNII \cite{huang2021unignn} achieves a 64-layer deep TNN by generalizing over-smoothing solutions from GNNs \cite{chen2020simple} to the HG domain. HSNs \cite{hajij2022high} generalize skip connections to allow signal to propagate further, but are still implemented as shallow networks.

 

\section{Conclusion}
In this work, we have provided a comprehensive, intuitive and critical view of the advances in TNNs through unifying notations and graphical illustrations. We have characterized each neural network by its choice of data domain and its model, which we further specify through choice of neighboring structure(s) and message-passing scheme. We hope that this review will make this rich body of work more accessible to practitioners whose fields would benefit from topology-sensitive deep learning.

\section*{Acknowledgments}
This work was supported by the National Science Foundation Grant Number 2134241.

\section{References Section}
\bibliographystyle{IEEEtran}
\bibliography{bibliography}

\vfill

\end{document}